%% file: iclr2027_conference.tex
\documentclass{article} 
\usepackage{iclr2027_conference,times}

\input{math_commands.tex}

\usepackage{graphicx}
\usepackage{booktabs}
\usepackage{algorithm}
\usepackage{algpseudocode}
\usepackage{multirow}
\usepackage{amsmath,amssymb}
\usepackage{xcolor}
\usepackage{colortbl}
\usepackage{caption}
\usepackage{wrapfig}
\usepackage{hyperref}
\usepackage{url}
\usepackage{subcaption}
\usepackage{tabularx}
\usepackage{array}
\newcommand{\method}{EditGRN}
\definecolor{papergreen}{RGB}{75,160,110}
\definecolor{paperblue}{RGB}{35,95,170}
\definecolor{paperorange}{RGB}{230,145,65}

\title{\method{}: Inversion-Free Image Editing by Source–Target Generative Synchronization}

\author{Jiahuan Wang, Jianan Chen, Rui Li, Guangming Lu \& Wenjie Pei \\
Harbin Institute of Technology, Shenzhen \\
\texttt{wangjiahuanszhit@163.com} \\
\texttt{lrhit@stu.hit.edu.cn} \\
\texttt{luguangm@hit.edu.cn} \\
\texttt{wenjiecoder@outlook.com}
\And
Jian Han \\
Tsinghua University \\
}

\newcommand{\meanerr}[2]{%
  \begin{tabular}[c]{@{}c@{}}
    $#1$\\[-3pt]
    {\tiny$\pm$\textnormal{#2}}
  \end{tabular}%
}
\iclrfinalcopy 
\begin{document}

\maketitle

\vspace{-1.0em}
\begin{center}
    \includegraphics[width=1\textwidth]{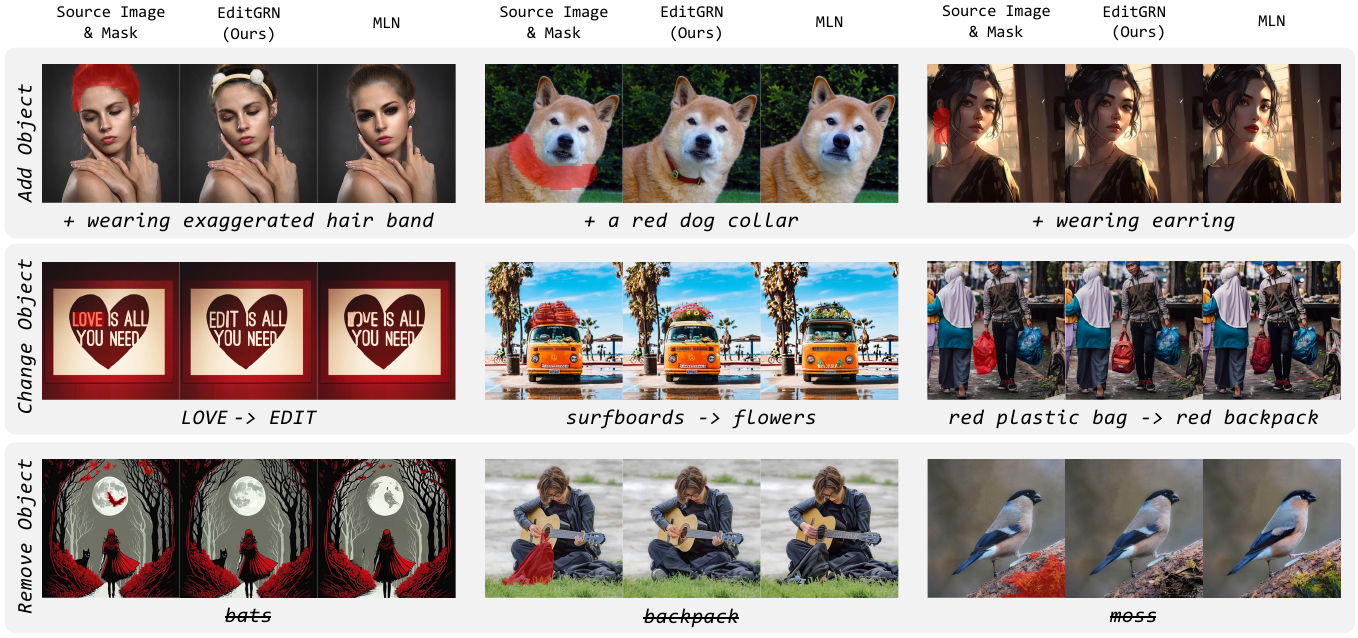}
    \captionof{figure}{Across object addition, change, and removal, our method, \textit{\method{}}, produces accurate, contextually coherent edits while preserving unedited content, whereas the state-of-the-art training- and inversion-free discrete method MLN~\cite{elghoussani2026mln} exhibits incomplete modifications and unintended appearance changes in these examples. Red masks indicate editing regions.}
\label{fig:teaser}
\end{center}

\begin{abstract}
Text-guided image editing faces a persistent trade-off between preserving source content and realizing coherent target semantics. We explore this challenge through training- and inversion-free image editing based on discrete generative models. Existing inversion-free discrete editors primarily rely on post-hoc probability correction, creating a disconnect between local token-level adjustments and globally coherent semantic generation. To bridge this gap, we introduce \textit{\method{}}, a training- and inversion-free framework with trajectory-wide source-target generative synchronization in discrete space. \textit{\method{}} adopts a dual-branch design---the source branch constructs source references and the target branch generates the target image---enabling trajectory-wide synchronization between the two. Exploiting GRN's ability to recover clean tokens from arbitrary binary corruptions, the source branch directly constructs corrupted source states at each refinement step to extract useful editing features, without recovering a specific source-generation trajectory. The target branch synchronizes with the source branch through the same corruption schedule, fusing the corresponding source context with editable representations intra-model. By synchronizing source context with target generation along the entire trajectory, our approach faithfully preserves the background while producing high-quality edits that blend seamlessly with the surrounding scene through intra-model fusion. PIE-Bench show state-of-the-art performance on background preservation and text alignment across diverse editing tasks against continuous and discrete baselines.
\end{abstract}

\section{Introduction}
Text-guided image editing requires balancing target realization against background preservation. Training-based editors require costly paired editing data~\citep{brooks2023instructpix2pix,zhang2023magicbrush,fu2024mgie,mu2025editar}, motivating training-free editing on pretrained generators~\citep{hertz2023prompt,cao2023masactrl}. Discrete models offer a distinct advantage: each token is either retained as source ground truth or independently resampled, providing a directly manipulable interface free from diffusion's entangled per-dimension Gaussian noise.

Image inversion traces generative dynamics backward from a source image to recover a noise trajectory that initializes target-conditioned generation~\citep{ju2024pnp,mokady2023null,wang2025rfsolver}. However, recent work shows that explicit inversion is not always essential for effective editing~\citep{xu2024infedit,kulikov2025flowedit,wang2025aredit,elghoussani2026mln}, shifting focus to incorporating source information directly into editing without inversion. 

In discrete generation, inversion-free editors such as AREdit ~\citep{wang2025aredit} and MLN~\citep{elghoussani2026mln} regulate tokens through post-hoc probability correction. Such output-level adjustment leaves a gap between local token nudges and coherent structure generation. As shown in Figure~\ref{fig:teaser}, it either fails to realize the target edit or entangles source and target features (\textit{e.g.}, the flowers and red backpack). This motivates injecting source context into the model's internal computation rather than correcting probabilities after prediction. Internal source-feature extraction has proven effective in continuous flow-based editors such as KV-Edit~\citep{zhu2025kvedit}, but obtaining these features relies on a specific and sequential source inversion trajectory. Direct Gaussian noising avoids inversion, yet the resulting states need not lie on a coherent source trajectory and therefore may not provide effective editing features. This raises a central question: can a discrete generative model provide useful internal source features without recovering a specific generation trajectory.

We propose \textit{\method{}}, a training- and inversion-free discrete framework built on trajectory-wide source-target generative synchronization. We build on Generative Refinement Networks (GRN)~\citep{han2026grn}, a discrete generative model trained to recover clean tokens from arbitrary binary-corrupted states rather than trajectory-consistent inputs. Leveraging this property, the source branch directly applies binary corruption to the clean source tokens at each refinement step, avoiding the recovery of a specific generation trajectory and the accumulation of sequential prediction errors. The source and edit branches share the same random corruption schedule, establishing source–target synchronization between the extracted source features and the corresponding edit representations at each refinement step, thereby enabling effective target editing. Inside the model, this synchronized source context fuses with editable-region representations, simultaneously preserving background fidelity, generating high-quality target content, and yielding smooth edited–unedited integration. Since the edit branch reasons only over editable tokens, its attention cost scales with the edit region rather than the full image. Smaller edits therefore run faster, and the extracted source features can be reused across multiple edits of the same image. On PIE-Bench, \textit{EditGRN} achieves state-of-the-art background preservation and text alignment. Our contributions are:
\begin{itemize}
    \item We present \textit{EditGRN}, a training- and inversion-free discrete editing framework built on trajectory-wide source–target synchronization: rather than correcting token probabilities post-hoc, source context is aligned with editing representations throughout refinement and fused intra-model.
    \item We observe and exploit GRN’s robustness to arbitrary binary corruption, while sharing the same random corruption schedule between the source and edit branches to obtain synchronized source features that effectively guide target editing without recovering a specific, coherent source-generation trajectory.
    \item Extensive experiments on PIE-Bench demonstrate that \textit{EditGRN} achieves state-of-the-art background preservation and text alignment, with computation that adaptively scales with edit-region size.
\end{itemize}

\section{Related Work}

\textbf{Training-based and training-free image editing.}
Training-based editors~\citep{brooks2023instructpix2pix,zhang2023magicbrush,fu2024mgie,mu2025editar} learn image-conditioned generation from costly paired or synthetic editing data.
Alternatively, pretrained generators can be adapted for editing without task-specific training.
Some methods optimize image-specific conditioning embeddings to improve reconstruction~\citep{mokady2023null}, while others operate directly at inference through attention control, feature sharing, or trajectory manipulation~\citep{hertz2023prompt,cao2023masactrl,kulikov2025flowedit,wang2025aredit}.
Our method requires neither task-specific training nor per-image optimization, using only forward inference with pretrained models.

\textbf{Inversion-based and inversion-free image editing.}
Inversion-based methods recover a source noise or latent trajectory before reusing its attention or features during target generation~\citep{hertz2023prompt,ju2024pnp,wang2025rfsolver,zhu2025kvedit}.
Inversion-free alternatives show that source-consistent editing can instead be constructed directly during inference: InfEdit couples source and target diffusion processes~\citep{xu2024infedit}, while FlowEdit defines a direct source-to-target flow~\citep{kulikov2025flowedit}.
Our method directly re-corrupts encoded source tokens to each GRN refinement level, avoiding recovery of a source trajectory or a continuous noise endpoint.

\textbf{Discrete generative models for image editing.}
Discrete editors adapt different generation paradigms: EditAR learns next-token editing~\citep{mu2025editar}, whereas DICE, VARIN, and EditInfinity recover source representations through discrete inversion or per-image optimization~\citep{he2026dice,dao2025varin,wang2025editinfinity}.
Inversion-free methods instead exploit source--target probability differences, as in AREdit~\citep{wang2025aredit}, or bias target logits toward source tokens, as in Masked Logit Nudging~\citep{elghoussani2026mln}.
Our approach centers on constructing source reference features without inversion.
Leveraging GRN's training over random binary-token corruption~\citep{han2026grn}, we directly corrupt the encoded source at each refinement level, rather than reconstructing its original generation history.

\section{Preliminaries: Generative Refinement Networks}
\label{sec:background}

\textbf{Visual Tokenizer.} GRN employs a visual tokenizer composed of an encoder $\mathcal{E}$, Hierarchical Binary Quantization (HBQ), and a decoder $\mathcal{D}$. Given an normalized encoder feature $F\in(-1,1)$, HBQ performs $M$ rounds hierarchical binary quantization:
\begin{equation}
    c_i=\sum_{j=1}^{i-1}\frac{\delta(q_j)}{2^j},
    \qquad
    q_i=
    \begin{cases}
        0, & F\leq c_i,\\
        1, & F>c_i,
    \end{cases}
    \label{eq:hbq}
\end{equation}
where $\delta(0)=-1$ and $\delta(1)=1$. The quantized feature is reconstructed as
$\widehat F=\sum_{i=1}^{M}\delta(q_i)2^{-i}$. Concatenating the binary decisions across feature channels gives the token map
\begin{equation}
    z
    =
    \mathcal{Q}_{\mathrm{HBQ}}\!\left(\mathcal{E}(x)\right)
    \in\{0,1\}^{D\times H\times W},
    \label{eq:grn_tokens}
\end{equation}
which can be mapped back to image space by $\mathcal{D}$.

\textbf{Generative Network.} During training, GRN learns to predict the ground-truth token map from a randomly corrupted input:
\begin{equation}
    \widetilde z
    =
    m\odot z+(1-m)\odot z_{\mathrm{rand}},
    \label{eq:grn_corruption}
\end{equation}
where $m\in\{0,1\}^{D\times H\times W}$ selects ground-truth $z$ or uniformly sampled random bits $z_{rand}$ at each entry. Given $\widetilde z$, text condition $c$, and retention ratio $p=\frac{\lVert m\rVert_1}{DHW}$, the Transformer predicts token logits
$\ell=f_\theta(\widetilde z,c,p)$, from which a token map is sampled. The model is optimized with the cross-entropy
\begin{equation}
    \mathcal{L}_{\mathrm{GRN}}
    =
    -\frac{1}{DHW}
    \sum_{k,d}
    \log p_\theta
    \left(z_{k,d}\mid\widetilde z,c,p\right).
    \label{eq:grn_loss}
\end{equation}
By training across randomly sampled retention ratio $p$, retention maps $m$, and replacement random binary tokens $z_{rand}$, GRN learns to recover tokens directly from arbitrary corrupted states, not just from trajectory-consistent inputs. At inference, GRN progressively refines a fully random binary token map by increasing its retention ratio from $0$ to $1$.

\begin{figure}[t]
    \centering
    \includegraphics[width=\linewidth]{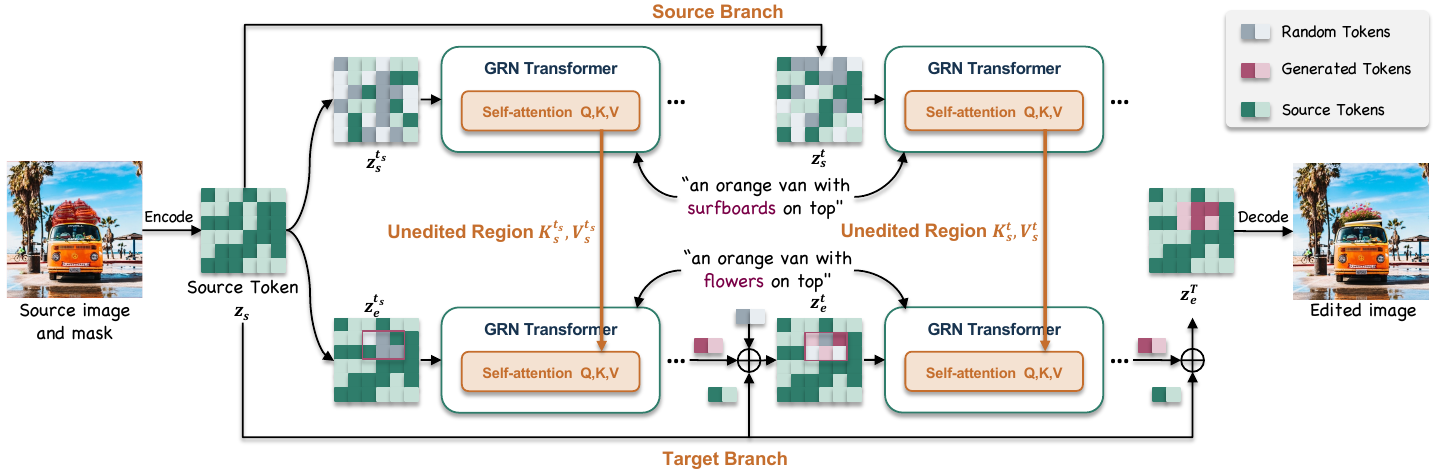}
    \caption{\textbf{Overview of \method{}}. The source branch (top) corrupts clean source tokens at each refinement stage and extracts unedited-region KV; the target branch (bottom) shares the same corruption schedule and replaces its unedited-region KV with the source ones inside every Transformer layer, generating only within the edit region while protected tokens remain anchored to the source.}
    \label{fig:overview}
\end{figure}

\section{Method}
\label{sec:method}
\subsection{Dual-branch Synchronized Editing Framework}
Given a source image $x_s$, source and target prompts $c_s$ and $c_e$, and a user-provided mask $R$, our goal is to modify tokens within $R$ while preserving those in the protected region $\bar{R}=1-R$. As shown in Figure~\ref{fig:overview}, EditGRN adopts a dual-branch design: the \textit{source branch} constructs source references, and the \textit{target branch} generates the target image. The two branches are synchronized under the same GRN refinement schedule, including retention maps $\{m_t\}_{t=t_s}^T$ and random binary tokens $z_{\mathrm{rand}}$, so that corruption intensity is aligned at every step. At each step, the source branch processes a corrupted source state with full self-attention and extracts source keys and values from the visual tokens. Inside every Transformer layer, the target branch replaces its protected region keys and values with these source keys and values, so target queries attend to source-provided context. Predictions are written back only inside $R$, while protected tokens are restored from source tokens. Algorithm~\ref{alg:method} summarizes the procedure.

\subsection{Source Reference Construction}
\label{sec:source_trajectory}

\textbf{Trajectory-decoupled source representations.}
Our key observation is that GRN recovers clean tokens from arbitrary binary corruptions without requiring the corrupted states to form a continuous trajectory. We exploit this property by constructing each source state directly from the clean source tokens at its corresponding corruption level:
\begin{equation}
z_s^t = m_t \odot z_s + (1-m_t)\odot z_{\mathrm{rand}}, \qquad p_s^t = \frac{\|m_t\|_1}{DHW}, \qquad t=t_s,\ldots,T.
\end{equation}
Each $z_s^t$ is constructed directly from $z_s$, rather than propagated from $z_s^{t-1}$. The retention map $m_t$ only needs to match the scheduled ratio, without requiring a transition from the preceding retention map, while $z_{\mathrm{rand}}$ is sampled once and shared across steps. Thus, source reference construction does not require recovering a specific source-generation trajectory and does not accumulate source prediction errors.

In contrast, in continuous trajectory-based source feature extraction, source states are coupled through the learned velocity field:
$\frac{d\mathbf{x}_s^t}{dt}=v_\theta(\mathbf{x}_s^t,t,c_s),\mathbf{x}_s^t=\mathbf{x}_s^0+\int_0^t v_\theta(\mathbf{x}_s^\tau,\tau,c_s)\,d\tau .$ Although direct Gaussian noising can construct $\bar{\mathbf{x}}_s^t=(1-t)\mathbf{x}_s^0+t\boldsymbol{\epsilon}$, the resulting state is not guaranteed to lie on the same coherent source generation trajectory $\frac{d\bar{\mathbf{x}}_s^t}{dt}\not\equiv v_\theta(\bar{\mathbf{x}}_s^t,t,c_s)$. Therefore, direct state construction alone does not ensure that the extracted source features align with the corresponding target generation state or effectively guide editing.

\textbf{Source context extraction.}
We next extract source context from these trajectory-decoupled representations for use in the target branch. Motivated by the successful reuse of background KV for guidance in flow-based editing~\citep{zhu2025kvedit}, source keys modulate target-query attention over source positions, and source values supply source context for target generation. Specifically, GRN processes the joint sequence $[z_s^t; c_s; p_s^t]$ with full self-attention at each step $t$, projecting hidden states into queries, keys, and values. We cache the resulting source keys and values $\{K_s^t, V_s^t\}_{t=t_s}^T$ corresponding to the visual tokens for all steps. Because GRN is trained to recover clean tokens from binary-corrupted inputs, the features extracted from each corrupted source state contain the information needed to reconstruct the source. Sharing the same random corruption schedule synchronizes these source features with the corresponding target representations at every refinement step, allowing them to effectively guide target-region generation through intra-model fusion rather than post-hoc probability correction.

\begin{algorithm}[t]
\caption{\method{} with trajectory-wide source-target generative synchronization}
\label{alg:method}
\begin{algorithmic}[1]
\Require Source image $x_s$; source prompt $c_s$, target prompt $c_e$; user-provided mask $R$; GRN $f_\theta$; retention maps
$\{m_t\}_{t=t_s}^{T}$; random binary tokens
$z_{\mathrm{rand}}$
\State $z_s\gets\mathcal{Q}_{\mathrm{HBQ}}(\mathcal{E}(x_s))\in\{0,1\}^{D\times H \times W}$
\State $z_s^{t_s}\gets m_{t_s}\odot z_s+(1-m_{t_s})\odot z_{\mathrm{rand}}$
\State $\bar R\gets 1-R$
\Statex \textit{// Source branch}
\For{$t=t_s,\ldots,T$}
    \State $p_s^t\gets\frac{\sum_{i,d}m_{t,i,d}}{DHW}$ 
    \State $\{K_{s}^t,V_{s}^t\}\gets f_\theta(z_s^t,c_s,p_s^t)$
    \State \textbf{if} $t<T$ \textbf{then} $z_s^{t+1}\gets m_{t+1}\odot z_s+(1-m_{t+1})\odot z_{\mathrm{rand}}$
\EndFor
\Statex \textit{// Target branch}

\State $z_e^{t_s}[R]\gets z_s^{t_s}[R]$, $z_e^{t_s}[\bar R]\gets z_s[\bar R]$
\For{$t=t_s,\ldots,T$}
    \State $p_e^t\gets\frac{\sum_{i,d}R_{i,d}m_{t,i,d}}{\sum_{i,d}R_{i,d}}$
    \State $\widehat z_e^t[R]\gets f_\theta\!\left(z_e^t,c_e,p_e^t;\{K_{s}^t[\bar R],V_{s}^t[\bar R]\}\right)$
    \State \textbf{if} $t<T$ \textbf{then} $z_e^{t+1}[R]\gets m_{t+1}[R]\odot\widehat z_e^t[R]+(1-m_{t+1}[R])\odot z_{\mathrm{rand}}[R]$, $z_e^{t+1}[\bar R]\gets z_s[\bar R]$
\EndFor

\State $\widehat z_e^T[\bar R]\gets z_s[\bar R]$
\State \Return $\widehat x_e=\mathcal{D}(\widehat z_e^T)$
\end{algorithmic}
\end{algorithm}

\subsection{Trajectory-Wide Source–Target Generative Synchronization}
\label{sec:editing_refinement}
\textbf{Trajectory-Wide Synchronization Corruption Schedule.}
The target branch starts at the same $t_s$ as the source branch and shares the same retention map $m_t$ and random binary tokens $z_{\mathrm{rand}}$. To initialize its state, we take the clean source tokens $z_s$ in the protected region and apply the same corrupted source state $z_s^{t_s}$ inside the edit region:
\begin{equation}
z_e^{t_s}[R]= z_s^{t_s}[R], z_e^{t_s}[\bar R]= z_s[\bar R], \qquad p_e^t = \frac{\sum_{i,d} R_{i,d} m_{t,i,d}}{\sum_{i,d} R_{i,d}} \approx p_s^t,
\end{equation}
\begin{wrapfigure}{r}{0.54\textwidth}
  \centering
  \vspace{-0.5cm}
  \includegraphics[width=0.54\textwidth]{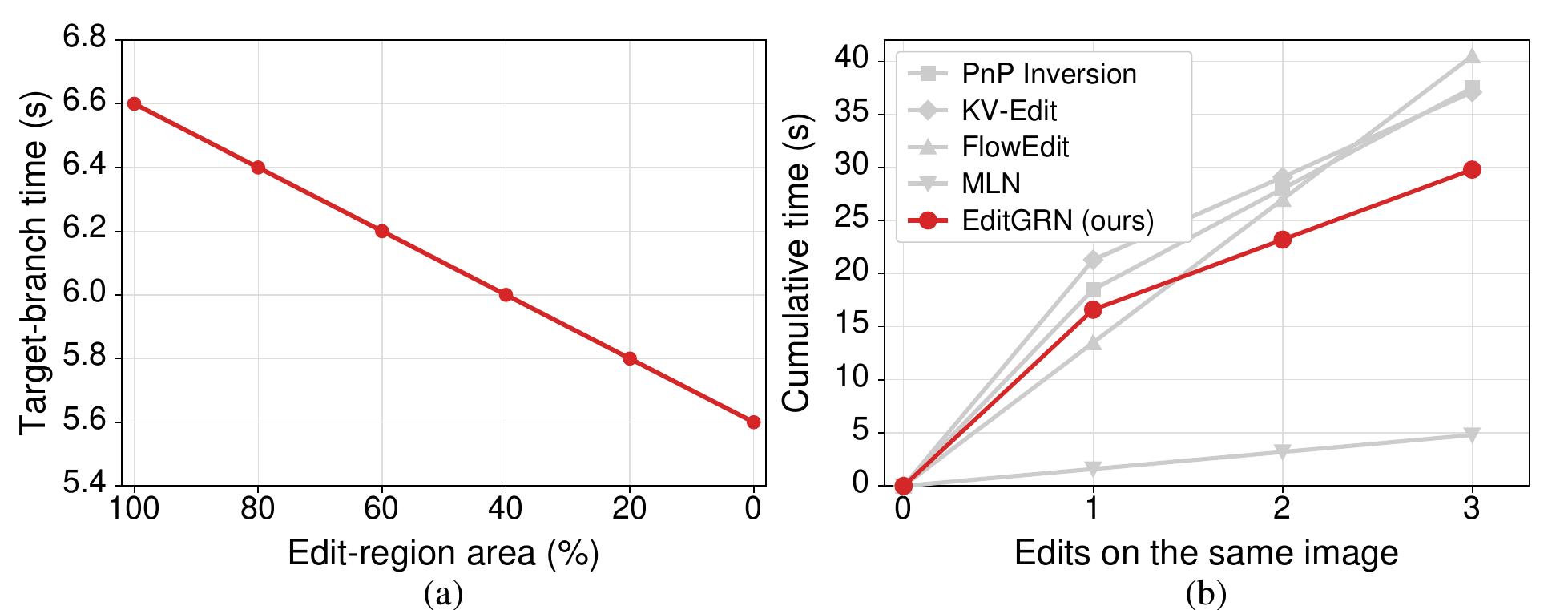}
  \caption{(a) Target-branch time scales with edit-region area. (b) Cumulative time for repeated edits. Measured on an NVIDIA L20 GPU.}
  \label{fig:time}
  \vspace{-0.6cm}
\end{wrapfigure}
where $\bar{R}=1-R$. Because $m_t$ selects bits uniformly across the token map, $p_e^t \approx p_s^t$, so the edit region experiences the same corruption intensity as the source branch at every step. This shared corruption schedule is the trajectory-wide synchronization: source context is always extracted at the corruption level matching the current target state, rather than from a fixed clean reference.

\paragraph{Intra-Model Region Fusion.}
At each Transformer layer, we replace the target branch's protected-region keys and values with the cached source ones at the same step:
$
\widetilde{K}_e^t = \mathrm{concat}[K_s^t[\bar{R}],\, K_e^t[R]], \qquad \widetilde{V}_e^t = \mathrm{concat}[V_s^t[\bar{R}],\, V_e^t[R]],
$
where $[R]$ and $[\bar{R}]$ index the edit region $R$ and protected region $\bar{R}$, respectively. Target queries in $R$ then attend over this fused visual context together with $c_e$ and $p_e^t$. After sampling the target prediction $\hat{z}_e^t[R]$ (defined only inside $R$), we corrupt the edit region for the next step while restoring protected tokens from the clean source:
\begin{equation}
z_e^{t+1}[R]= m_{t+1}[R]\odot\widehat z_e^t[R]+(1-m_{t+1}[R])\odot z_{\mathrm{rand}}[R], \qquad z_e^{t+1}[\bar R]= z_s[\bar R].
\end{equation}
By fusing synchronized source context with target representations inside the model at every step, the edit region generates with consistent background guidance while protected tokens remain.

\begin{table*}[t]
\caption{\textbf{Quantitative results on PIE-Bench.} Cont. and Disc. denote continuous (diffusion/flow) and discrete
generative models, respectively. Train indicates additional task-specific
training, Opt per-image optimization, and Inv explicit image inversion. For SGR-Edit, U and A
denote user-provided and attention-derived masks, respectively.
CLIP$_\mathrm{W}$ and CLIP$_\mathrm{E}$ measure alignment over the whole image
and edited region.}
\label{tab:main}
\centering
\small
\setlength{\tabcolsep}{1.5pt}
\renewcommand{\arraystretch}{1.06}
\begin{tabular*}{\textwidth}{@{\extracolsep{\fill}}lccccrrrrrr@{}}
\toprule
\multirow{2}{*}{Method} & \multirow{2}{*}{Type} &
\multirow{2}{*}{Train} & \multirow{2}{*}{Opt} & \multirow{2}{*}{Inv} &
\multicolumn{4}{c}{Background Preservation} &
\multicolumn{2}{c}{Text Alignment}\\
\cmidrule(lr){6-9}\cmidrule(lr){10-11}
& & & & & PSNR$\uparrow$ & LPIPS$_{\times10^3}\downarrow$ &
MSE$_{\times10^4}\downarrow$ & SSIM$_{\times10^2}\uparrow$ &
CLIP$_\mathrm{W}\uparrow$ & CLIP$_\mathrm{E}\uparrow$\\
\midrule
\multicolumn{11}{l}{\emph{Additional training or per-image optimization (Train=Y or Opt=Y)}}\\
InstructPix2Pix & Cont. & Y & N & N & 16.69 & 271.33 & 392.22 & 68.39 & 23.49 & 22.20\\
MGIE & Cont. & Y & N & N & 21.20 & 142.25 & 295.11 & 77.52 & 24.28 & 21.79\\
EditAR & Disc. & Y & N & N & 21.32 & 117.15 & 130.27 & 75.13 & 24.87 & 21.87\\
Null-text + P2P & Cont. & N & Y & Y & 27.03 & 60.67 & 35.86 & 84.11 & 24.75 & 21.86\\EditInfinity & Disc. & N & Y & Y & 27.95 & 33.08 & 24.27 & 92.12 & 26.41 & 23.47\\
\midrule
\multicolumn{11}{l}{\emph{Training-free: continuous generators (Train=N and Opt=N)}}\\
DDIM + P2P & Cont. & N & N & Y & 17.87 & 208.80 & 219.88 & 71.14 & 25.01 & 22.44\\
PnP Inversion & Cont. & N & N & Y & 22.46 & 106.06 & 80.45 & 79.68 & 25.41 & 22.62\\
InfEdit & Cont. & N & N & N & 28.51 & 47.58 & 32.09 & 85.66 & 25.03 & 22.22\\
RF-Edit & Cont. & N & N & Y & 24.83 & 113.15 & 52.46 & 83.38 & 25.57 & 22.26\\
KV-Edit & Cont. & N & N & Y & 35.87 & 9.92 & 4.69 & -- & 22.91 & --\\
FlowEdit & Cont. & N & N & N & 21.96 & 112.19 & 94.99 & 83.08 & 25.25 & 22.58\\
\midrule
\multicolumn{11}{l}{\emph{Training-free: discrete generators (Train=N and Opt=N)}}\\
DICE & Disc. & N & N & Y & 27.29 & 52.90 & 43.76 & 89.79 & 23.79 & 21.23\\
VARIN & Disc. & N & N & Y & 26.54 & 54.04 & 38.33 & 85.39 & 25.05 & 21.49\\
AREdit & Disc. & N & N & N & 24.19 & 87.00 & -- & 83.70 & 25.42 & 22.77\\
MLN & Disc. & N & N & N & 29.70 & 36.50 & 23.30 & 86.80 & 26.15 & 22.72\\
\midrule
\rowcolor{gray!12}
\textbf{\method{} (U)} & Disc. & N & N & N & \textbf{38.08} & \textbf{7.37} & \textbf{2.32} & \textbf{98.80} & \textbf{27.29} & \textbf{24.65}\\
\rowcolor{gray!12}
\textit{\method{} (A)} & Disc. & N & N & N & 33.22 & 16.87 & 15.07 & 97.59 & 26.80 & 23.83\\
\bottomrule
\end{tabular*}
\end{table*}

\begin{figure}[t]
    \centering
    \includegraphics[width=\linewidth]{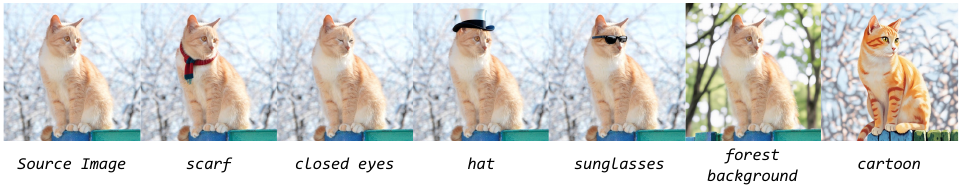}
    \caption{Multiple edits of the same image.}
    \label{fig:multi_edit}
\end{figure}

\textbf{Area-Adaptive Editing.}
The source branch attends over all $N_I$ tokens, while the target branch only computes visual queries for the $N_R$ tokens inside $R$, so the target-branch cost scales as $\mathcal{O}(N_R N_I)$ rather than $\mathcal{O}(N_I^2)$; as shown in Figure~\ref{fig:time} (a), target-branch time decreases linearly as the edit region shrinks. This advantage amplifies for repeated edits on the same image (Figure~\ref{fig:multi_edit}). Source features are extracted once and reused, whereas PnP re-inverts on every edit, KV-Edit pays a one-time but expensive inversion, and FlowEdit/MLN pay the full editing cost each time. EditGRN pays only once source once and then the target branch cost per edit, which results in a shallower cumulative-time slope (Figure~\ref{fig:time} (b)). While MLN is the fastest, it can result in fragmented, poorly integrated edits in Figure~\ref{fig:teaser}. EditGRN retains efficiency without sacrificing quality.

\section{Experiments}
\subsection{Experimental Setup}
\textbf{Comparison Methods.}
We compare three groups:
(1) editors requiring additional training or per-image optimization:
InstructPix2Pix~\citep{brooks2023instructpix2pix},
MGIE~\citep{fu2024mgie}, EditAR~\citep{mu2025editar},
Null-text + P2P~\citep{mokady2023null,hertz2023prompt},
and EditInfinity~\citep{wang2025editinfinity};
(2) training-free continuous editors:
DDIM + P2P~\citep{hertz2023prompt},
PnP Inversion~\citep{ju2024pnp}, InfEdit~\citep{xu2024infedit},
RF-Edit~\citep{wang2025rfsolver}, KV-Edit~\citep{zhu2025kvedit},
and FlowEdit~\citep{kulikov2025flowedit};
(3) training-free discrete editors:
DICE~\citep{he2026dice}, VARIN~\citep{dao2025varin},
AREdit~\citep{wang2025aredit}, and MLN~\citep{elghoussani2026mln}.

\textbf{Benchmark.}
PIE-Bench~\citep{ju2024pnp}, a widely adopted benchmark for text-guided image editing.
It covers nine editing tasks-object replacement, addition, deletion,
and changes to content, pose, color, material, background, and style.
The images span natural and artificial scenes across four categories:
animals, humans, indoor scenes, and outdoor scenes.
Each example provides a source image with its corresponding source
prompt, a target editing prompt, and an annotated editing mask.

\textbf{Metrics.}
Following PIE-Bench, we report background PSNR, LPIPS, MSE, and SSIM, and CLIP similarity for the whole image and edited region~\citep{caron2021dino,zhang2018lpips,wang2004ssim,radford2021clip}.

\textbf{Implementation Details.}
We use the pretrained GRN text-to-image model \footnote{\url{https://huggingface.co/bytedance-research/GRN/blob/main/GRN_T2I_2B_FSA_251600.pth}}. Experiments runs on a single NVIDIA L20 GPU. By default, we start at $t_s=14$ with $p_s^{t_s}=0.10$, retaining approximately $10\%$ of source bits and requiring only $T-t_s=36$ refinement steps out of $T=50$.

\subsection{Comparison to State-of-the-Arts}
\begin{wrapfigure}{R}{0.5\textwidth}
  \centering
  \vspace{-0.5cm}
  \includegraphics[width=0.5\textwidth]{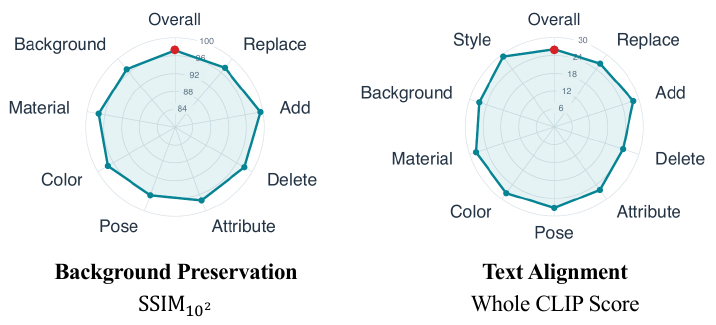}
  \caption{\textbf{Per-task results on PIE-Bench.} EditGRN's overall score (red dot) already exceeds other methods, and its per-task scores closely track this overall level across all categories.}
  \label{fig:radar_image}
  \vspace{-0.3cm}
\end{wrapfigure}
\textbf{Quantitative results.}
Under the PIE-Bench protocol, \method{} (U, User-Provided Mask) achieves the best results across all background-preservation and text-alignment, as shown in Table~\ref{tab:main}. These improvements hold over methods requiring training and training-free continuous editors. Among inversion-free discrete editors, \method{} also improves both aspects over AREdit and MLN, which employ prediction-probability guidance. Figure~\ref{fig:radar_image} further demonstrates strong performance across diverse editing tasks, rather than gains concentrated in a particular task category. Like KV-Edit and EditInfinity, \method{} (U) uses user-provided mask, while achieving stronger performance than them under the same evaluation protocol.

Beyond user-provided masks, we also evaluate \method{} (A, Attention-Derived Mask), which automatically localizes edits using attention maps. Following Prompt-to-Prompt~\citep{hertz2023prompt} and MLN~\cite{elghoussani2026mln}, we identify edited words from source-target prompt differences and extract visual-to-text attention maps from the source branch at $p_s^t=0.8$. We aggregate visual-to-text attention across subwords to localize edits, reusing the source branch forward pass. Otsu thresholding~\citep{otsu1979threshold} then converts the attention maps into binary masks separating editable and background regions. For object addition, we instead use the target prompt in the source branch to extract attention maps for newly added words. \method{} (A) still achieves competitive performance, suggesting limited reliance on precisely specified editing masks.

\begin{figure}[!htb]
    \centering
    \includegraphics[width=\linewidth]{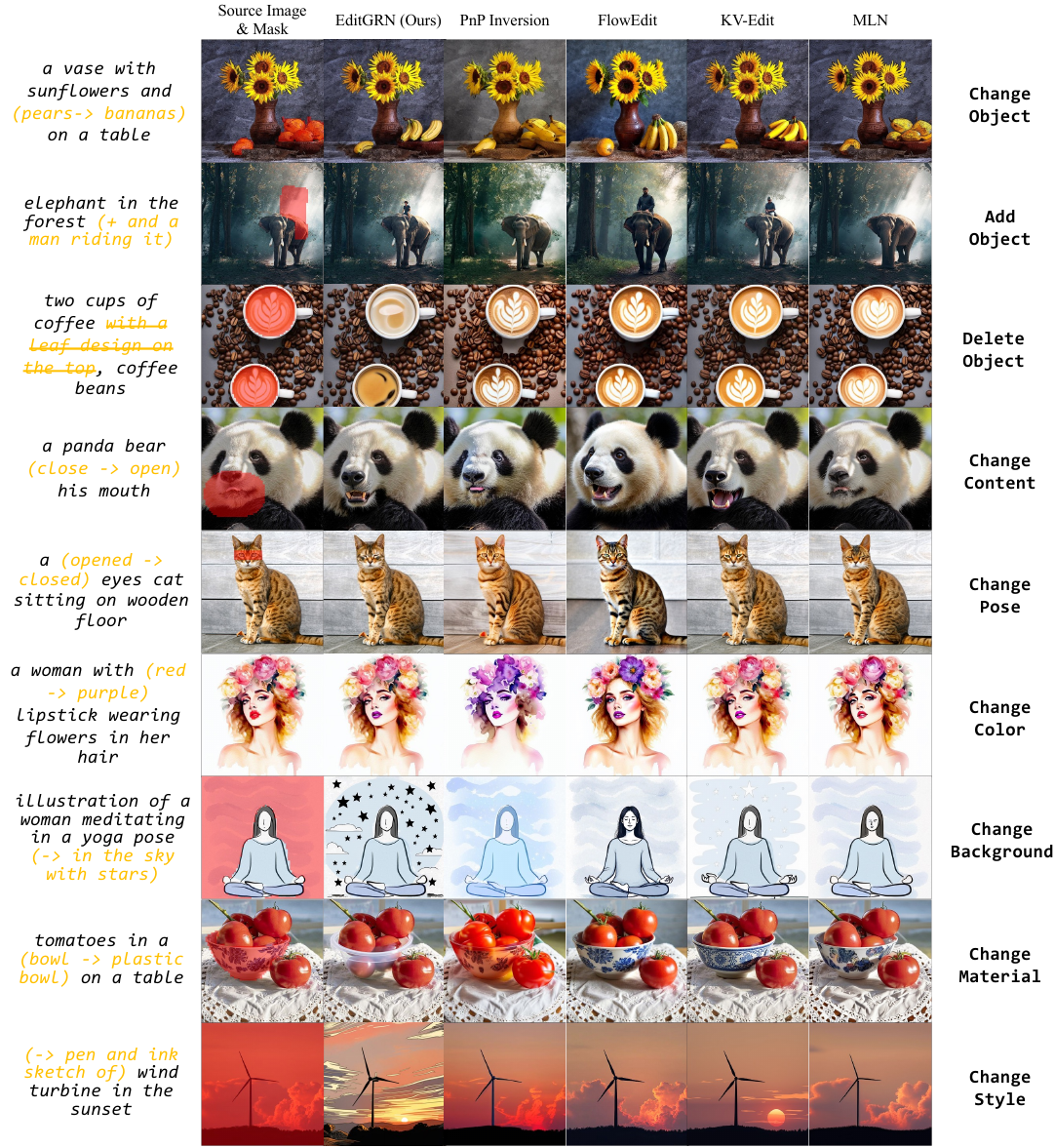}
    \caption{\textbf{Qualitative results on PIE-Bench across all nine tasks.} The red mask denotes the edited region $R$, expected to follow the target prompt, while other regions retain the background.}
    \label{fig:compare_all}
\end{figure}

\textbf{Qualitative Results.}
We compare across all PIE-Bench editing tasks in Figure~\ref{fig:compare_all}. For space, we include four representative training-free editors: PnP Inversion, an classical inversion-based diffusion method; FlowEdit, an classical inversion-free flow method; KV-Edit, a flow-based inversion method that reuses source background KV; and MLN, a state-of-the-art inversion-free discrete editor. Across diverse tasks, including complex multi-object real-world scenes and text editing in Figure~\ref{fig:teaser}, \method{} preserves unedited content while producing target-aligned modifications.

\subsection{Ablation Studies}
\label{sec:ablations}
Ablation studies are performed on the ‘random class’ of PIE-Bench, covering all types of editing and allowing an efficient and unbiased evaluation.

\begin{table*}[!htb]
\caption{Quantitative results of ablating on Randomness Robustness and Synchronized Corruption Schedule.}
\label{tab:ablation_random}
\centering
\small
\setlength{\tabcolsep}{3.5pt}
\renewcommand{\arraystretch}{1.08}
\begin{tabular*}{\textwidth}{@{\extracolsep{\fill}}lcccccc@{}}
\toprule
\multirow{2}{*}{Variant} &
\multicolumn{4}{c}{Background Fidelity} &
\multicolumn{2}{c}{Text Alignment}\\
\cmidrule(lr){2-5}\cmidrule(lr){6-7}
& PSNR$\uparrow$
& LPIPS$_{\times 10^3}\downarrow$
& MSE$_{\times 10^4}\downarrow$
& SSIM$_{\times 10^2}\uparrow$
& CLIP$_{\mathrm{W}}\uparrow$
& CLIP$_{\mathrm{E}}\uparrow$\\
\midrule
\rowcolor{gray!12}
\method{}
& \meanerr{39.00}{1e-2}
& \meanerr{6.78}{1e-4}
& \meanerr{2.06}{3e-6}
& \meanerr{98.90}{9e-5}
& \meanerr{26.39}{0.05}
& \meanerr{23.31}{0.08}\\
Unsynchronized $\{m_t\}$
& 38.94 &  6.69 & 2.04 &  98.90 &  26.34 &  23.18\\
Unsynchronized $z_{rand}$
& 38.91 &  6.68 &  2.04 &  98.90 &  26.20 & 23.33\\
Unsynchronized $\{m_t\}, z_{rand}$
& 38.80 & 6.74 & 2.01 & 98.90 & 25.96 & 23.02\\
Unsynchronized Corruption Level
& 38.30 & 7.63 & 2.15 & 98.84 & 23.31 & 19.17\\
\bottomrule
\end{tabular*}
\end{table*}

\begin{table*}[!htb]
\caption{Quantitative results of ablating on Protected Region Writeback and Source Context.}
\label{tab:ablation_KV_bg}
\centering
\small
\setlength{\tabcolsep}{3.5pt}
\renewcommand{\arraystretch}{1.08}
\begin{tabular*}{\textwidth}{@{\extracolsep{\fill}}lcccccc@{}}
\toprule
\multirow{2}{*}{Variant} &
\multicolumn{4}{c}{Background Fidelity} &
\multicolumn{2}{c}{Text Alignment}\\
\cmidrule(lr){2-5}\cmidrule(lr){6-7}
& PSNR$\uparrow$
& LPIPS$_{\times 10^3}\downarrow$
& MSE$_{\times 10^4}\downarrow$
& SSIM$_{\times 10^2}\uparrow$
& CLIP$_{\mathrm{W}}\uparrow$
& CLIP$_{\mathrm{E}}\uparrow$\\
\midrule
\rowcolor{gray!12}
\method{}
& 39.00
& 6.78
& 2.06
& 98.90
& 26.39
& 23.31\\
$\,-$ Protected-region writeback
& 38.29 & 11.64 & 2.69 & 98.56 & 26.36 & 23.32\\
$\,-$ Source KV 
& 38.63 & 6.56 & 2.06 & 98.89 & 24.72 & 21.00\\
 $\,-$ Both
&  15.30 &  298.47 &  434.32 &  62.57 &  26.67 &  23.09 \\
\bottomrule
\end{tabular*}
\end{table*}

\begin{table*}[!htb]
\caption{Quantitative results of ablating on Trajectory-Decoupled Source in GRN and Flow.}
\label{tab:ablation_KV_Edit}
\centering
\small
\setlength{\tabcolsep}{3.5pt}
\renewcommand{\arraystretch}{1.08}
\begin{tabular*}{\textwidth}{@{\extracolsep{\fill}}lcccccc@{}}
\toprule
\multirow{2}{*}{Variant} &
\multicolumn{4}{c}{Background Fidelity} &
\multicolumn{2}{c}{Text Alignment}\\
\cmidrule(lr){2-5}\cmidrule(lr){6-7}
& PSNR$\uparrow$
& LPIPS$_{\times 10^3}\downarrow$
& MSE$_{\times 10^4}\downarrow$
& SSIM$_{\times 10^2}\uparrow$
& CLIP$_{\mathrm{W}}\uparrow$
& CLIP$_{\mathrm{E}}\uparrow$\\
\midrule
\rowcolor{gray!12}
\method{}
& 39.00
& 6.78
& 2.06
& 98.90
& 26.39
& 23.31\\
KV-Edit w/ inversion & 35.81 &  8.69
 & 5.30 & 95.74 &  25.05 &  22.72 \\
KV-Edit w/o inversion & 34.71 & 10.73 & 6.27 & 95.72 & 24.56 & 22.23 \\
\bottomrule
\end{tabular*}
\end{table*}

\begin{wrapfigure}{r}{0.6\textwidth}
  \centering
  \vspace{-0.4cm}
  \includegraphics[width=0.6\textwidth]{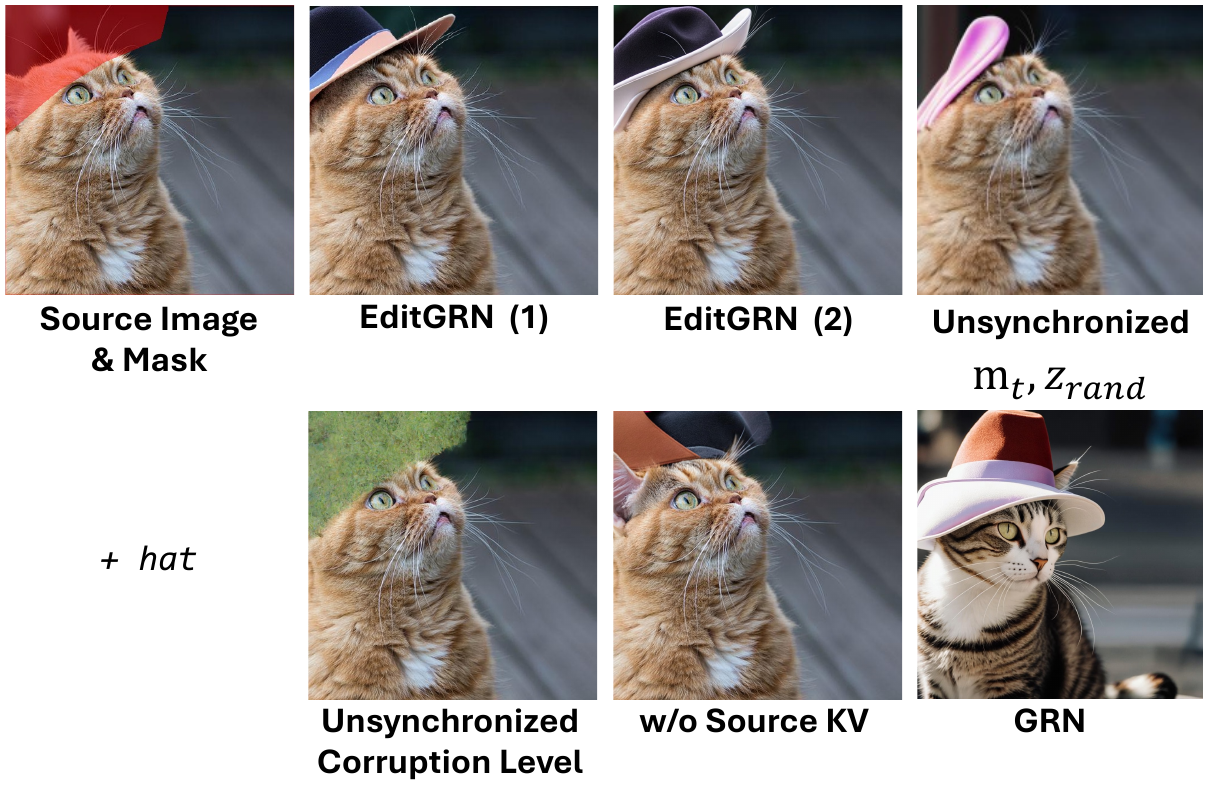}
  \caption{Qualitative results of ablation studies.}
  \label{fig:ablation}
  \vspace{-0.5cm}
\end{wrapfigure}

\textbf{Ablation on Randomness Robustness.}
We run EditGRN with shared $\{m_t\}$ and $z_{\mathrm{rand}}$ across the two branches but vary the random seeds across independent runs. The first line of Table~\ref{tab:ablation_random} shows that metrics remain stable, and the two runs produce visually consistent edits (Figure~\ref{fig:ablation}, EditGRN (1--2)). This supports our trajectory-decoupled design. Since GRN is trained to recover clean tokens from arbitrary corrupted states, directly corrupting the clean source with different random seeds can still provide source KVs that consistently guide target editing. This suggests that useful editing features can be obtained without recovering source-specific generation randomness.


\textbf{Ablation on Synchronized Corruption Schedule.}
We progressively disrupt synchronization by independently sampling $m_t$, $z_{\mathrm{rand}}$, or both across the source and target branches. Table~\ref{tab:ablation_random} shows that changing either component alone has a limited effect, while changing both causes a larger degradation. Figure~\ref{fig:ablation} further shows that independently sampling both $m_t$ and $z_{\mathrm{rand}}$ may reduce the coherence between the edited hat and the surrounding fur.

For unsynchronized corruption level, we reuse KVs from the final, least-corrupted source state throughout refinement, fixing the source corruption level at its minimum while the target corruption level progressively decreases, leaving them mismatched at every step except the last. Consequently, even KVs extracted from the least-corrupted source state provide less effective guidance when their corruption level is mismatched with that of the target state, resulting in the largest degradation in Table~\ref{tab:ablation_random}. Figure~\ref{fig:ablation} also shows that this mismatch substantially weakens the editing result.


\textbf{Ablation on Protected Region Writeback and Source Context.}
To assess the role of Protected Region Writeback, we construct a variant in which the target branch predicts and refines the entire token map rather than only the edit region. In this variant, the protected tokens are neither initialized nor restored from the clean source tokens, while synchronized source KV guidance remains unchanged. The detailed procedure is provided in Algorithm \ref{alg:without_bg}. As shown by the `$\text{-}$ Protected-region writeback' variant in Table~\ref{tab:ablation_KV_bg}, background fidelity deteriorates. Source KV supplies contextual features but does not enforce equality between protected and source tokens. Token writeback explicitly enforces this constraint, improving background preservation without compromising text alignment.

To assess the role of Source Context, we remove source KV injection while retaining the remaining components. Since no source KV are required in this variant, the source branch is removed entirely. As shown by the `$\text{-}$ Source KV' variant in Table~\ref{tab:ablation_KV_bg}, both text-alignment metrics decrease substantially. Figure~\ref{fig:ablation} further shows that source KV provides contextual cues from the source image, helping generate the edited object consistently with the surrounding content.

Finally, removing both Protected Region Writeback and Source KV Injection reduces the method to bare GRN generation initialized from the corrupted source state $\mathbf z_s^{t_s}$. Without explicit background preservation or source context guidance, the model regenerates the full token map under the target prompt. GRN can still generate the target semantics, but cannot preserve and localize the edit to the source image without the two proposed components. Consequently, CLIP-W remains high, while background fidelity degrades substantially. This confirms that the gains do not come from the backbone alone.

\begin{wrapfigure}{r}{0.6\textwidth}
  \centering
  \vspace{-0.5cm}
  \includegraphics[width=0.6\textwidth]{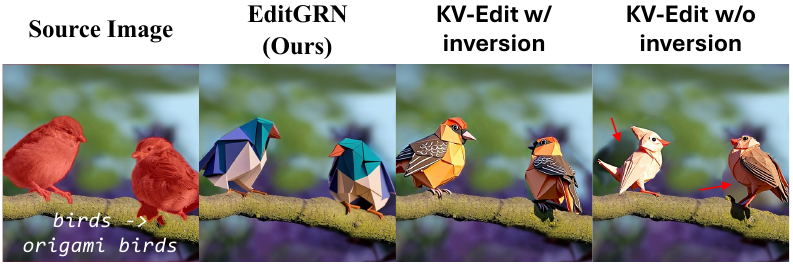}
  \caption{Qualitative results of ablation studies on Trajectory-Decoupled Source in GRN and Flow.}
  \label{fig:flow_noise_qual}
  \vspace{-0.4cm}
\end{wrapfigure}

\textbf{Ablation on Trajectory-Decoupled Source in GRN and Flow.}
To assess whether trajectory-decoupled source-reference construction can provide useful KVs for editing without inversion, we contrast EditGRN’s direct binary corruption $ \mathbf z_s^{t_s} = \mathbf m_{t_s}\odot\mathbf z_s+ (1-\mathbf m_{t_s})\odot\mathbf z_{\mathrm{rand}}$ with KV-Edit’s ~\citep{zhu2025kvedit} sequential RF inversion $ \mathbf{x}_s^t=\mathbf{x}_s^0+\int_0^t v_\theta(\mathbf{x}_s^\tau,\tau,c_s)\,d\tau$ and its inversion-free variant based on direct Gaussian noising $ \bar{\mathbf{x}}_s^t = (1-t)\mathbf{x}_s^0+t\boldsymbol{\epsilon},  \boldsymbol{\epsilon}\sim\mathcal N(\mathbf 0,\mathbf I).$ Here, $\mathbf{x}_s^0$ denotes the clean source latent, and $v_\theta$ is the RF velocity field conditioned on the source prompt $c_s$. The two KV-Edit variants share the same flow-based target branch and background-KV injection, differing only in their source-reference construction.

As shown in Table ~\ref{tab:ablation_KV_Edit}, KV-Edit with inversion consistently outperforms its direct-noising variant. This indicates that KV-Edit extracts more effective source KVs from trajectory-consistent inversion states than from directly Gaussian-noised states. In contrast, EditGRN obtains strong performance without inversion, suggesting that GRN’s arbitrary binary-corruption training enables directly constructed source states to provide useful, synchronized editing features. Figure~\ref{fig:flow_noise_qual} provides consistent qualitative evidence. The direct-noising flow variant introduces artifacts when changing the birds into origami, whereas EditGRN successfully performs these edits.

\textbf{Ablation on Initial Retention Ratio.}
The initial retention ratio $p_s^{t_s}$ determines how much source information is retained in the edit region at the start of refinement: a larger $p_s^{t_s}$ preserves more source content but leaves less room for target realization.  Table~\ref{tab:ablation_ts} and Figure~\ref{fig:ablation_ts} in the Appendix show that this trade-off manifests as improved background fidelity at the cost of weaker text alignment. We choose $p_s^{t_s}=0.10$ with $t_s=14$ to balance both objectives.

\section{Conclusion}
We presented EditGRN, a training- and inversion-free framework for text-guided image editing in discrete space. Our key finding is that GRN's ability to recover clean tokens from arbitrary binary corruptions enables the direct extraction of effective source features without recovering a specific source-generation trajectory. EditGRN exploits this property through synchronized source and target branches, using a shared corruption schedule to provide effective source guidance for target editing through intra-model fusion. Experiments demonstrate strong background preservation and text alignment, highlighting trajectory-decoupled source-feature extraction as a promising direction for coherent and efficient image editing.

\subsection*{AI use statement}
In this work, we used generative AI tools to assist with understanding existing code, diagnosing and fixing implementation bugs, and improving the clarity and grammar of the manuscript. All AI-assisted code changes were manually reviewed and validated through experiments, and all AI-assisted writing was checked and revised by the authors. Generative AI was not used to fabricate data, results, citations, or scientific claims. We take responsibility for the final content of this work, including text, claims or artifacts produced with the aid of generative AI.
\bibliography{iclr2027_conference}
\bibliographystyle{iclr2027_conference}

\clearpage
\appendix
\section{Appendix}

\subsection{Implementation Details for the Protected-Region Writeback Ablation}
Our algorithm \ref{alg:method} initializes the protected region with source tokens: $z_e^{t_s}[R]\gets z_s^{t_s}[R], z_e^{t_s}[\bar R]\gets z_s[\bar R]$. Then, the model predicts only the edit region $R$ tokens $\widehat z_e^t[R]\gets f_\theta\!\left(z_e^t,c_e,p_e^t;\{K_{s}^t[\bar R],V_{s}^t[\bar R]\}\right)$, after which the protected tokens are explicitly restored at every step: $z_e^{t+1}[R]\gets m_{t+1}[R]\odot\widehat z_e^t[R]+(1-m_{t+1}[R])\odot z_{\mathrm{rand}}[R], z_e^{t+1}[\bar R]\gets z_s[\bar R]$.

In the without Protected Region Writeback ablation, we (i) remove the initialization of the protected region with clean source tokens by replacing Algorithm \ref{alg:method} line 9 with $z_e^{t_s}\gets z_s^{t_s}$ ; (ii) replace Algorithm \ref{alg:method} line 13 with $\widehat z_e^t\gets f_\theta\!\left(z_e^t,c_e,p_e^t;\{K_s^t[\bar R],V_s^t[\bar R]\}\right)$ so that the model predicts the entire token map;  and (iii) replace Algorithm \ref{alg:method} line 14 with $\mathbf{z}_{e}^{t+1} \leftarrow m_{t+1} \odot \hat{\mathbf{z}}_{e}^{t} + \left(1-m_{t+1}\right)\odot \mathbf{z}_{\mathrm{rand}}$ thereby removing the stepwise restoration of the protected region.

Thus, the protected tokens are also predicted and corrupted rather than kept fixed. Source-KV injection alone cannot enforce exact token preservation, explaining the degraded background metrics.

\begin{algorithm}[h]
\caption{\method{} without protected-region writeback}
\label{alg:without_bg}
\begin{algorithmic}[1]
\Require Source image $x_s$; source prompt $c_s$, target prompt $c_e$;
user-provided mask $R$; GRN $f_\theta$; retention maps
$\{m_t\}_{t=t_s}^{T}$; random binary tokens $z_{\mathrm{rand}}$

\State $z_s\gets\mathcal{Q}_{\mathrm{HBQ}}(\mathcal{E}(x_s))
\in\{0,1\}^{D\times H\times W}$
\State $z_s^{t_s}\gets
m_{t_s}\odot z_s+(1-m_{t_s})\odot z_{\mathrm{rand}}$
\State $\bar R\gets 1-R$

\Statex \textit{// Source branch}
\For{$t=t_s,\ldots,T$}
    \State $p_s^t\gets\frac{\sum_{i,d}m_{t,i,d}}{DHW}$
    \State $\{K_s^t,V_s^t\}\gets f_\theta(z_s^t,c_s,p_s^t)$
    \State \textbf{if} $t<T$ \textbf{then}
    $z_s^{t+1}\gets
    m_{t+1}\odot z_s+(1-m_{t+1})\odot z_{\mathrm{rand}}$
\EndFor

\Statex \textit{// Target branch without protected-region writeback}
\State {$z_e^{t_s}\gets z_s^{t_s}$}

\For{$t=t_s,\ldots,T$}
    \State $p_e^t\gets
    \frac{\sum_{i,d}m_{t,i,d}}{DHW}$
    \State {$\widehat z_e^t\gets
    f_\theta\!\left(
    z_e^t,c_e,p_e^t;
    \{K_s^t[\bar R],V_s^t[\bar R]\}
    \right)$}
    \State \textbf{if} $t<T$ \textbf{then}
    {$z_e^{t+1}\gets
    m_{t+1}\odot\widehat z_e^t+
    (1-m_{t+1})\odot z_{\mathrm{rand}}$}
\EndFor

\State \Return $\widehat x_e=\mathcal{D}(\widehat z_e^T)$
\end{algorithmic}
\end{algorithm}

\subsection{Extended Analysis of $p_s^{t_s}$ and $t_s$}
\begin{figure}[h]
    \centering
    \includegraphics[width=\linewidth]{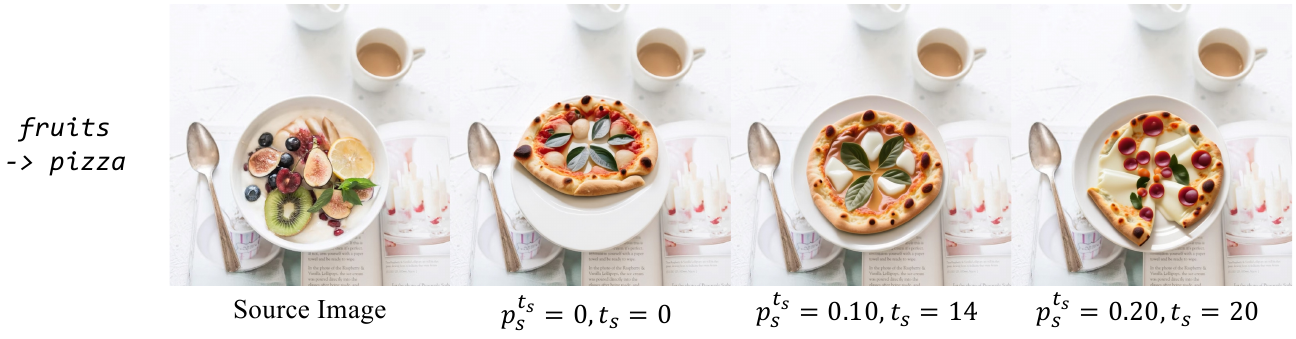}
    \caption{Qualitative comparison of different initial retention ratio and start step.}
    \label{fig:ablation_ts}
\end{figure}

The trade-off in Table~\ref{tab:ablation_ts} arises because $p_s^{t_s}$ directly determines the initial composition of the edit region: when $p_s^{t_s}$ is small, the edit region begins closer to pure random tokens, giving the model full freedom to generate target content but with weaker constraints on how the new object should integrate with the surrounding source scene. As $p_s^{t_s}$ grows, more source tokens are pinned in place, which anchors the edit region to the source layout but also injects source structure that may conflict with the target prompt. Figure~\ref{fig:ablation_ts} illustrates this: with $p_s^{t_s}=0.20$, the generated pizza bears source-like artifacts that should have been replaced, while with $p_s^{t_s}=0$ the pizza is cleaner but its shadow and lighting blend less naturally with the table setting. The start step $t_s$ acts on the same axis: starting refinement later (larger $t_s$) means the source branch and target branch share fewer, less-corrupted steps, reducing synchronization benefit. Our default $p_s^{t_s}=0.10$ and $t_s=14$ gives the balance between target realization and scene integration.

\begin{table}[h]
\centering
\caption{Ablation on initial retention ratio $p_s^{t_s}$ and start step $t_s$.}
\label{tab:ablation_ts}
\begin{tabular}{lccccccc}
\toprule
 & & \multicolumn{4}{c}{Background Fidelity} & \multicolumn{2}{c}{Text Alignment} \\
\cmidrule(lr){3-6} \cmidrule(lr){7-8}
$p_s^{t_s}$ & $t_s$ & PSNR$\uparrow$ & LPIPS$_{\times 10^3}\downarrow$ & MSE$_{\times 10^4}\downarrow$ & SSIM$_{\times 10^2}\uparrow$ & CLIP$_{\mathrm{W}}\uparrow$ & CLIP$_{\mathrm{E}}\uparrow$ \\
\midrule
0.00 & 0  & 38.49 & 7.30 & 2.30 & 98.86 & 26.59 & 23.67 \\
0.10 & 14 & 39.00 & 6.78 & 2.06 & 98.90 & 26.39 & 23.31 \\
0.20 & 20 & 39.72 & 6.26 & 1.80 & 98.95 & 25.83 & 22.88 \\
\bottomrule
\end{tabular}
\end{table}

\begin{table*}[h]
\caption{Sources of the baseline results reported in Table~\ref{tab:main}.}
\label{tab:baseline_sources}
\centering
\small
\setlength{\tabcolsep}{4pt}
\renewcommand{\arraystretch}{1.12}
\begin{tabularx}{\textwidth}{
    >{\raggedright\arraybackslash}p{0.25\textwidth}
    >{\raggedright\arraybackslash}X
    >{\centering\arraybackslash}p{0.17\textwidth}}
\toprule
\textbf{Comparison Methods} &
\textbf{Source Paper} &
\textbf{Venue} \\
\midrule
InstructPix2Pix, MGIE, EditAR &
\textit{EditAR: Unified Conditional Generation with Autoregressive Models} &
CVPR 2025 \\

Null-text + P2P, DDIM + P2P, PnP Inversion &
\textit{PnP Inversion: Boosting Diffusion-based Editing with 3 Lines of Code} &
ICLR 2024 \\

EditInfinity &
\textit{EditInfinity: Image Editing with Binary-Quantized Generative Models} &
NeurIPS 2025 \\

InfEdit &
\textit{Inversion-Free Image Editing with Language-Guided Diffusion Models} &
CVPR 2024 \\

RF-Edit, FlowEdit &
\textit{DNAEdit: Direct Noise Alignment for Text-Guided Rectified Flow Editing} &
NeurIPS 2025 Spotlight \\

KV-Edit, excluding style change &
\textit{KV-Edit: Training-Free Image Editing for Precise Background Preservation} &
ICCV 2025 \\

KV-Edit, style-change subset &
Rerun using the official KV-Edit implementation &
--- \\

DICE, VARIN &
\textit{Discrete Noise Inversion for Next-scale Autoregressive Text-based Image Editing} &
ECCV 2026 \\

AREdit &
\textit{Training-Free Text-Guided Image Editing with Visual Autoregressive Model} &
ICCV 2025 \\

MLN &
\textit{Prompt-Guided Image Editing with Masked Logit Nudging in Visual Autoregressive Models} &
CVPR 2026 \\
\bottomrule
\end{tabularx}
\end{table*}

\begin{figure}[h]
    \centering
    \includegraphics[width=\linewidth]{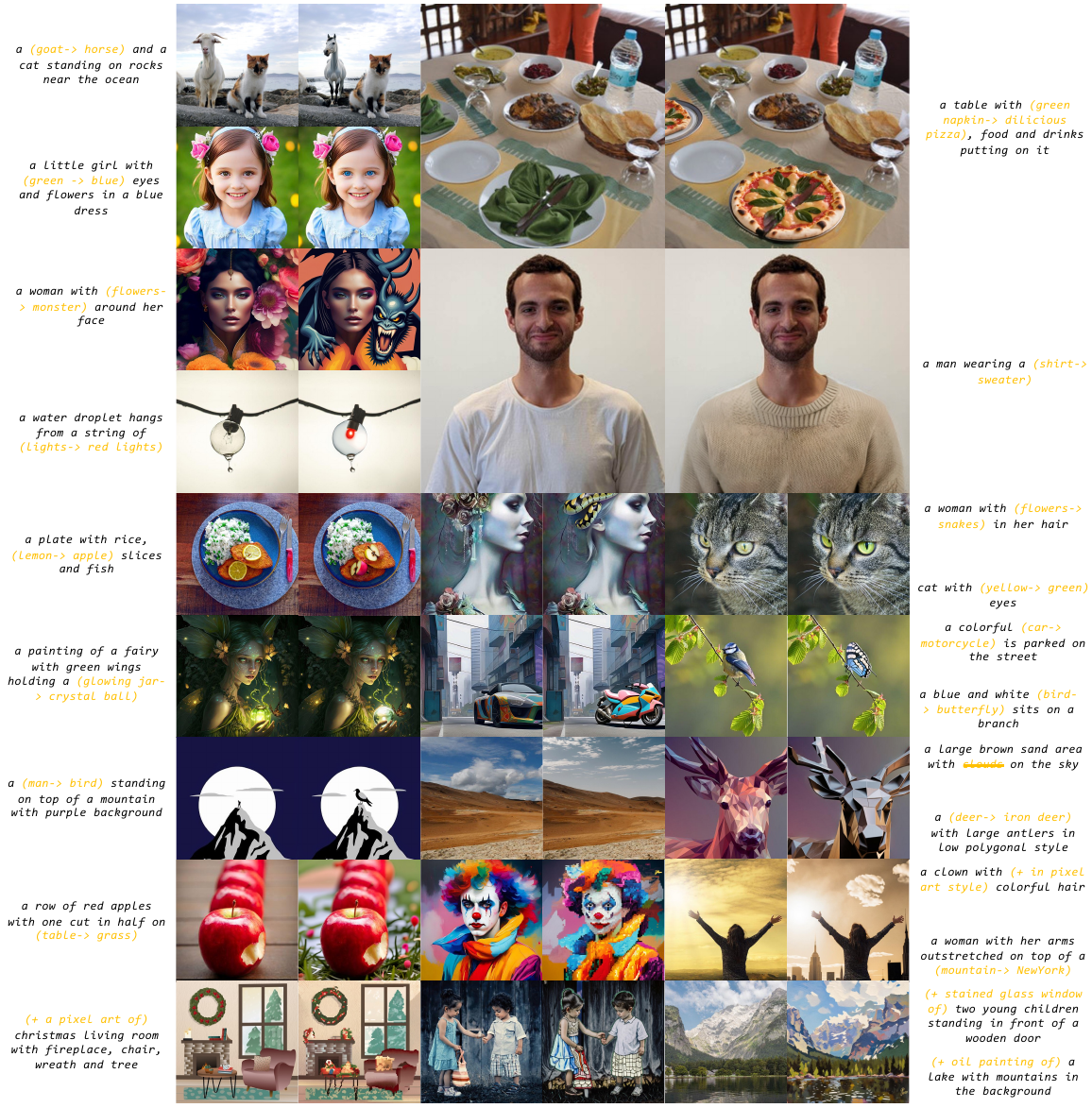}
    \caption{\textbf{Additional editing results.} For each pair, the left is the source image and the right is EditGRN's output. The zoomed-in views highlight smooth boundary blending between edited and preserved regions.}
    \label{fig:more_results}
\end{figure}

\subsection{Sources of Baseline Results}
\label{app:baseline_sources}
Except for the missing style-change subset of KV-Edit, which we reran, all results of comparison methods in Table~\ref{tab:baseline_sources} are taken from previously published papers. Since KV-Edit excludes style-change tasks from its original PIE-Bench evaluation, We reran KV-Edit on the style-change subset of PIE-Bench using its official implementation \url{https://github.com/Xilluill/KV-Edit}, and combined these results with its originally reported scores. Therefore, the completed KV-Edit results cover the same PIE-Bench tasks as EditGRN and all comparison methods.

\subsection{Additional Qualitative Results}
Figure~\ref{fig:more_results} presents more editing results across diverse tasks. EditGRN successfully realizes the target edits while faithfully preserving the unedited regions. As highlighted in the zoomed-in views, the boundaries between edited and preserved regions blend naturally, with consistent lighting, texture, and shadow-confirming the effectiveness of intra-model source context fusion.

\subsection{Limitations}
This work focuses on static image editing; its extension to video and temporal consistency remain unexplored. Applying our method independently to each frame does not explicitly enforce cross-frame consistency and may introduce appearance variations or temporal flickering. Motion and occlusion further alter the editing region and its surrounding context, making frame-wise source guidance insufficient to guarantee temporally coherent edits. Future work should investigate temporally consistent source-feature construction and edit-region tracking, while assessing the computational and memory costs of caching features across frames.

\end{document}

%% file: math_commands.tex
\usepackage{amsmath,amsfonts,bm}

\def\eqref#1{equation~\ref{#1}}

\def\1{\bm{1}}

\DeclareMathAlphabet{\mathsfit}{\encodingdefault}{\sfdefault}{m}{sl}
\SetMathAlphabet{\mathsfit}{bold}{\encodingdefault}{\sfdefault}{bx}{n}



%% file: iclr2027_conference.bib
@article{han2026grn,
  title={Generative Refinement Networks for Visual Synthesis},
  author={Han, Jian and Liu, Jinlai and Wang, Jiahuan and Peng, Bingyue and Yuan, Zehuan},
  journal={arXiv preprint arXiv:2604.13030},
  year={2026}
}

@inproceedings{wang2025editinfinity,  
    title={EditInfinity: Image Editing with Binary-Quantized Generative Models},  
    author={Wang, Jiahuan and Chen, Yuxin and Yu, Jun and Lu, Guangming and Pei, Wenjie},  booktitle={Advances in Neural Information Processing Systems},  
    year={2025}
}

@inproceedings{hertz2023prompt,
  title={Prompt-to-Prompt Image Editing with Cross-Attention Control},
  author={Hertz, Amir and Mokady, Ron and Tenenbaum, Jay and Aberman, Kfir and Pritch, Yael and Cohen-Or, Daniel},
  booktitle={International Conference on Learning Representations},
  year={2023}
}

@inproceedings{mokady2023null,
  title={Null-text Inversion for Editing Real Images Using Guided Diffusion Models},
  author={Mokady, Ron and Hertz, Amir and Aberman, Kfir and Pritch, Yael and Cohen-Or, Daniel},
  booktitle={IEEE/CVF Conference on Computer Vision and Pattern Recognition},
  year={2023}
}

@inproceedings{ju2024pnp,
  title={PnP Inversion: Boosting Diffusion-based Editing with 3 Lines of Code},
  author={Ju, Xuan and Zeng, Ailing and Bian, Yuxuan and Liu, Shaoteng and Xu, Qiang},
  booktitle={International Conference on Learning Representations},
  year={2024}
}

@inproceedings{cao2023masactrl,
  title={MasaCtrl: Tuning-Free Mutual Self-Attention Control for Consistent Image Synthesis and Editing},
  author={Cao, Mingdeng and Wang, Xintao and Qi, Zhongang and Shan, Ying and Qie, Xiaohu and Zheng, Yinqiang},
  booktitle={IEEE/CVF International Conference on Computer Vision},
  year={2023}
}

@inproceedings{xu2024infedit,
  title={Inversion-Free Image Editing with Language-Guided Diffusion Models},
  author={Xu, Sihan and Huang, Yidong and Pan, Jiayi and Ma, Ziqiao and Chai, Joyce},
  booktitle={IEEE/CVF Conference on Computer Vision and Pattern Recognition},
  year={2024}
}

@inproceedings{kulikov2025flowedit,
  title={FlowEdit: Inversion-Free Text-Based Editing Using Pre-Trained Flow Models},
  author={Kulikov, Vladimir and Kleiner, Matan and Huberman-Spiegelglas, Inbar and Michaeli, Tomer},
  booktitle={IEEE/CVF International Conference on Computer Vision},
  pages={19721--19730},
  year={2025}
}

@inproceedings{wang2025rfsolver,
  title={Taming Rectified Flow for Inversion and Editing},
  author={Wang, Jiangshan and Pu, Junfu and Qi, Zhongang and Guo, Jiayi and Ma, Yue and Huang, Nisha and Chen, Yuxin and Li, Xiu and Shan, Ying},
  booktitle={International Conference on Machine Learning},
  pages={64044--64058},
  year={2025}
}

@inproceedings{zhu2025kvedit,
  title={KV-Edit: Training-Free Image Editing for Precise Background Preservation},
  author={Zhu, Tianrui and Zhang, Shiyi and Shao, Jiawei and Tang, Yansong},
  booktitle={IEEE/CVF International Conference on Computer Vision},
  year={2025}
}

@inproceedings{wang2025aredit,
  title={Training-Free Text-Guided Image Editing with Visual Autoregressive Model},
  author={Wang, Yufei and Guo, Lanqing and Li, Zhihao and Huang, Jiaxing and Wang, Pichao and Wen, Bihan and Wang, Jian},
  booktitle={IEEE/CVF International Conference on Computer Vision},
  year={2025}
}

@inproceedings{brooks2023instructpix2pix,
  title={InstructPix2Pix: Learning to Follow Image Editing Instructions},
  author={Brooks, Tim and Holynski, Aleksander and Efros, Alexei A.},
  booktitle={IEEE/CVF Conference on Computer Vision and Pattern Recognition},
  pages={18392--18402},
  year={2023}
}

@inproceedings{zhang2023magicbrush,
  title={MagicBrush: A Manually Annotated Dataset for Instruction-Guided Image Editing},
  author={Zhang, Kai and Mo, Lingbo and Chen, Wenhu and Sun, Huan and Su, Yu},
  booktitle={Advances in Neural Information Processing Systems, Datasets and Benchmarks Track},
  volume={36},
  year={2023}
}

@inproceedings{fu2024mgie,
  title={Guiding Instruction-Based Image Editing via Multimodal Large Language Models},
  author={Fu, Tsu-Jui and Hu, Wenze and Du, Xianzhi and Wang, William Yang and Yang, Yinfei and Gan, Zhe},
  booktitle={International Conference on Learning Representations},
  year={2024}
}

@inproceedings{mu2025editar,
  title={EditAR: Unified Conditional Generation with Autoregressive Models},
  author={Mu, Jiteng and Vasconcelos, Nuno and Wang, Xiaolong},
  booktitle={IEEE/CVF Conference on Computer Vision and Pattern Recognition},
  pages={7899--7909},
  year={2025}
}

@article{otsu1979threshold,
  title={A Threshold Selection Method from Gray-Level Histograms},
  author={Otsu, Nobuyuki},
  journal={IEEE Transactions on Systems, Man, and Cybernetics},
  volume={9},
  number={1},
  pages={62--66},
  year={1979},
  doi={10.1109/TSMC.1979.4310076}
}

@inproceedings{he2026dice,
  title={DICE: Discrete Inversion Enabling Controllable Editing for Masked Generative Models},
  author={He, Xiaoxiao and Dao, Quan and Han, Ligong and Wen, Song and Bai, Minhao and Liu, Di and Zhang, Han and Juefei-Xu, Felix and Tan, Chaowei and Liu, Bo and Min, Martin Renqiang and Li, Kang and Ahmed, Faez and Srivastava, Akash and Li, Hongdong and Huang, Junzhou and Metaxas, Dimitris N.},
  booktitle={IEEE/CVF Winter Conference on Applications of Computer Vision},
  pages={762--772},
  year={2026}
}

@article{dao2025varin,
  title={Discrete Noise Inversion for Next-Scale Autoregressive Text-Based Image Editing},
  author={Dao, Quan and He, Xiaoxiao and Han, Ligong and Nguyen, Ngan Hoai and Nobari, Amin Heyrani and Ahmed, Faez and Zhang, Han and Nguyen, Viet Anh and Metaxas, Dimitris N.},
  journal={arXiv preprint arXiv:2509.01984},
  year={2025}
}

@inproceedings{elghoussani2026mln,
  title={Prompt-Guided Image Editing with Masked Logit Nudging in Visual Autoregressive Models},
  author={El-Ghoussani, Amir and H{\"o}lle, Marc and Carneiro, Gustavo and Belagiannis, Vasileios},
  booktitle={IEEE/CVF Conference on Computer Vision and Pattern Recognition Findings},
  pages={4810--4820},
  year={2026}
}

@inproceedings{caron2021dino,
  title={Emerging Properties in Self-Supervised Vision Transformers},
  author={Caron, Mathilde and Touvron, Hugo and Misra, Ishan and J{\'e}gou, Herv{\'e} and Mairal, Julien and Bojanowski, Piotr and Joulin, Armand},
  booktitle={IEEE/CVF International Conference on Computer Vision},
  year={2021}
}

@inproceedings{zhang2018lpips,
  title={The Unreasonable Effectiveness of Deep Features as a Perceptual Metric},
  author={Zhang, Richard and Isola, Phillip and Efros, Alexei A. and Shechtman, Eli and Wang, Oliver},
  booktitle={IEEE/CVF Conference on Computer Vision and Pattern Recognition},
  year={2018}
}

@article{wang2004ssim,
  title={Image Quality Assessment: From Error Visibility to Structural Similarity},
  author={Wang, Zhou and Bovik, Alan C. and Sheikh, Hamid R. and Simoncelli, Eero P.},
  journal={IEEE Transactions on Image Processing},
  volume={13}, number={4}, pages={600--612}, year={2004}
}

@inproceedings{radford2021clip,
  title={Learning Transferable Visual Models From Natural Language Supervision},
  author={Radford, Alec and Kim, Jong Wook and Hallacy, Chris and Ramesh, Aditya and Goh, Gabriel and Agarwal, Sandhini and Sastry, Girish and Askell, Amanda and Mishkin, Pamela and Clark, Jack and others},
  booktitle={International Conference on Machine Learning},
  year={2021}
}
